\documentclass[conference]{IEEEtran}
\IEEEoverridecommandlockouts

\usepackage{cite}
\usepackage{amsmath,amssymb,amsfonts}
\usepackage{algorithmic}
\usepackage{graphicx}
\usepackage{textcomp}
\usepackage{xcolor}
\usepackage{booktabs}
\usepackage{orcidlink}
\def\BibTeX{{\rm B\kern-.05em{\sc i\kern-.025em b}\kern-.08em
    T\kern-.1667em\lower.7ex\hbox{E}\kern-.125emX}}

\begin{document}

\title{Distributed Kalman--Consensus Filtering with Adaptive Uncertainty Weighting for Multi-Object Tracking in Mobile Robot Networks}
\author{
\IEEEauthorblockN{1st\textsuperscript{st} Niusha Khosravi\orcidlink{0009-0002-8938-0996}}
\IEEEauthorblockA{\textit{Instituto Superior Técnico} \\
\textit{University of Lisbon} \\
Lisbon, Portugal \\
niusha.khosravi@tecnico.ulisboa.pt}
\and
\IEEEauthorblockN{2nd\textsuperscript{nd} Rodrigo Ventura\orcidlink{0000-0002-5655-9562}}
\IEEEauthorblockA{\textit{Instituto Superior Técnico} \\
\textit{University of Lisbon} \\
Lisbon, Portugal \\
rodrigo.ventura@tecnico.ulisboa.pt}
\and
\IEEEauthorblockN{3\textsuperscript{rd} Meysam Basiri\orcidlink{0000-0002-8456-6284}}
\IEEEauthorblockA{\textit{Instituto Superior Técnico} \\
\textit{University of Lisbon} \\
Lisbon, Portugal \\
meysam.basiri@tecnico.ulisboa.pt}
}

\maketitle

\begin{abstract}
This paper presents an implementation and evaluation of a Distributed Kalman--Consensus Filter (DKCF) for Multi-Object Tracking (MOT) in mobile robot networks operating under partial observability and heterogeneous localization uncertainty. A key challenge in such systems is the fusion of information from agents with differing localization quality, where frame misalignment can lead to inconsistent estimates, track duplication, and ghost tracks. To address this issue, we build upon the MOTLEE framework and retain its frame-alignment methodology, which uses consistently tracked dynamic objects as transient landmarks to improve relative pose estimates between robots. On top of this framework, we propose an uncertainty-aware adaptive consensus weighting mechanism that dynamically adjusts the influence of neighbor information based on the covariance of the transmitted estimates, thereby reducing the impact of unreliable data during distributed fusion. Local tracking is performed using a Kalman Filter (KF) with a Constant Velocity Model (CVM) and Global Nearest Neighbor (GNN) data association. simulation  results demonstrate that adaptive weighting effectively protects local estimates from inconsistent data, yielding a MOTA improvement of 0.09 for agents suffering from localization drift, although system performance remains constrained by communication latency.
\end{abstract}


\section{Introduction}
Autonomous robots operating in dynamic environments must detect and track multiple moving objects (DATMO) while maintaining situational awareness and navigating safely. Multi-object tracking (MOT) enables robots to predict the motion of obstacles and other agents, plan collision-free paths, and coordinate effectively with teammates. The problem becomes particularly challenging when the observer itself is mobile, the environment is only partially known, and sensor data are corrupted by noise and occlusions.

Traditional MOT approaches typically employ a single robot equipped with exteroceptive sensors such as lidar or cameras~\cite{wang2002simultaneous,wang2003online,wang2007simultaneous}. These systems integrate motion compensation, feature extraction, clustering, data association, and state estimation into a tracking-by-detection pipeline. Despite substantial progress, single-robot systems remain fundamentally limited by restricted sensor fields of view, occlusion effects, and local localization errors~\cite{choi2012general}.

Collaborative multi-robot systems can overcome these limitations by exploiting spatially distributed sensing and computation. Through information sharing, robots can jointly track objects that are intermittently visible to individual robots, and improve system robustness against sensor failures and occlusions. Distributed estimation methods, such as Distributed Kalman--Consensus Filters (DKCF), provide a principled approach to fuse local estimates into consistent global estimates without requiring a central fusion node~\cite{Peterson2023,tian2022dl}.

Recent work by Peterson \textit{et al.}~\cite{Peterson2023} introduced MOTLEE, a distributed mobile multi-object tracking framework with localization error elimination. The MOTLEE approach combines distributed Kalman consensus filtering with principled frame alignment strategies to address the challenge of tracking in environments where localization uncertainty is significant. In this work, we adopt the MOTLEE framework and extend it with an adaptive uncertainty weighting scheme that dynamically weights the contribution of each robot's local estimate based on its localization uncertainty. This extension addresses a key limitation of standard consensus approaches: they typically assign equal or static weights to all agents, regardless of heterogeneous localization quality across the robot network.

The main contributions of this paper are:
\begin{itemize}
    \item \textbf{System Implementation:} A complete ROS-based distributed tracking pipeline built upon the MOTLEE framework, integrating laser-based DBSCAN detection, Kalman Filter tracking, and consensus fusion.
    \item \textbf{Adaptive Weighting Integration:} The implementation and analysis of an uncertainty-aware weighting scheme that dynamically adjusts consensus gains based on the estimated quality of neighbor tracks.
\end{itemize}

The remainder of this paper is organized as follows. Section~\ref{sec:related} reviews related work on MOT with uncertain localization and distributed filtering. Section~\ref{sec:local} presents the local detection and tracking framework. Section~\ref{sec:dkcf} describes the DKCF-based global tracking scheme and the proposed adaptive weighting mechanism. Section~\ref{sec:results} reports simulation results, and Section~\ref{sec:discussion} provides discussion and conclusions.

\section{Related Work}
\label{sec:related}
Multi-object tracking has been extensively studied in computer vision and robotics communities, often formulated as tracking-by-detection approaches that associate object detections across time to recover trajectories~\cite{ref1}. Research in this area has focused on improving data association algorithms, sensor fusion strategies, and robustness to occlusion and clutter.

\subsection{Single-View Multi-Object Tracking with Uncertain Localization}
When robot localization is uncertain, multi-object tracking becomes tightly coupled with simultaneous localization and mapping (SLAM). Wang \textit{et al.}~\cite{ref33} formulated the joint SLAM and MOT problem, explicitly distinguishing between static landmarks and dynamic objects in a unified probabilistic framework. More recent approaches leverage factor graphs~\cite{ref37} and use tracked objects to refine visual SLAM estimates~\cite{ref38}, thereby improving localization robustness under motion and occlusion.

\subsection{Multi-Robot Multi-Object Tracking with Uncertain Localization}
Multi-robot MOT with uncertain localization has received comparatively less research attention. Lin \textit{et al.}~\cite{lin2004exact} demonstrated that estimates from multiple static sensors can jointly estimate dynamic biases, though their method is tailored to range-and-bearing sensors and does not address mobile platforms. Moratuwage \textit{et al.}~\cite{moratuwage2013collaborative} proposed a collaborative SLAM and tracking framework using random finite sets and the probability hypothesis density filter, but it requires centralized fusion of all sensor data. Ahmad \textit{et al.}~\cite{ahmad2013cooperative} formulated joint cooperative MOT and localization as pose-graph optimization in robotic soccer contexts, though assuming known data associations.

Peterson \textit{et al.}~\cite{Peterson2023} recently introduced MOTLEE, a distributed approach specifically designed for mobile multi-object tracking with localization error elimination. MOTLEE employs distributed Kalman-consensus filtering combined with principled frame alignment strategies to handle localization uncertainty. The present work adopts the core MOTLEE consensus framework and extends it with adaptive weighting to account for heterogeneous localization uncertainties across the robot network.

\subsection{Adaptive and Weighted Consensus Filtering}
Distributed consensus algorithms have been extensively studied in control and estimation literature~\cite{olfati2007consensus}. When agent estimates have heterogeneous reliabilities, weighted consensus approaches that dynamically adjust agent weights based on estimate uncertainty become advantageous. Our adaptive weighting approach builds upon this principle, applying uncertainty-aware weighting specifically to the context of distributed multi-robot tracking.

\section{Local Detection and Tracking}
\label{sec:local}
This section describes the local perception pipeline executed independently on each robot: object detection and clustering from laser scans, and multi-object tracking using a Kalman Filter with GNN-based data association.

\subsection{Object Detection and Clustering}
Each robot is equipped with a 2D lidar that provides range--bearing measurements $(r_i, \theta_i)$. Points are first converted to Cartesian coordinates in the laser frame:
\begin{equation}
x_i = r_i \cos \theta_i, \quad y_i = r_i \sin \theta_i.
\end{equation}
Using the robot pose obtained from the TF tree, points are transformed into the odometry (global) frame to ensure temporal consistency across scans.

To identify potential objects, we apply the Density-Based Spatial Clustering of Applications with Noise (DBSCAN) algorithm. DBSCAN groups points into clusters based on two parameters: the neighborhood radius $\epsilon$ and the minimum number of points $\text{MinPts}$ required to form a dense region. Clusters exceeding a predefined maximum size are discarded to reject large static structures such as walls, while remaining clusters are treated as candidate moving objects. For each cluster, we compute the centroid as the local measurement for tracking.

\subsection{Dynamic Model and Kalman Filter}
The state of each tracked object is modeled using a constant-velocity motion model in the plane:
\begin{equation}
\mathbf{x}(k) = [x(k), \dot{x}(k), y(k), \dot{y}(k)]^\top,
\end{equation}
with discrete-time linear dynamics:
\begin{equation}
\mathbf{x}(k+1) = \mathbf{F} \mathbf{x}(k) + \mathbf{w}(k),
\end{equation}
where the state transition matrix is:
\begin{equation}
\mathbf{F} = \begin{bmatrix}
1 & T & 0 & 0 \\
0 & 1 & 0 & 0 \\
0 & 0 & 1 & T \\
0 & 0 & 0 & 1
\end{bmatrix}
\end{equation}
and $T$ is the sampling period. The process noise satisfies $\mathbf{w}(k) \sim \mathcal{N}(0, \mathbf{Q})$ and accounts for unmodeled accelerations.

Each detection provides a position measurement:
\begin{equation}
\mathbf{z}(k) = [x(k), y(k)]^\top = \mathbf{H}\mathbf{x}(k) + \mathbf{v}(k),
\end{equation}
with measurement matrix:
\begin{equation}
\mathbf{H} = \begin{bmatrix}
1 & 0 & 0 & 0 \\
0 & 0 & 1 & 0
\end{bmatrix}
\end{equation}
and measurement noise $\mathbf{v}(k) \sim \mathcal{N}(0, \mathbf{R})$.

For each active track, standard Kalman Filter prediction and update equations propagate the state and covariance estimate. New tracks are initialized from unmatched detections with appropriately large initial covariance.

\subsection{Data Association with Global Nearest Neighbor and Hungarian Algorithm}
At each time step, the set of predicted track states is associated with the set of current detections using the Global Nearest Neighbor (GNN) approach. The association cost between track $i$ and detection $j$ is quantified by the Mahalanobis distance:
\begin{equation}
M^2_{ij} = (\mathbf{z}_j - \mathbf{H}\mathbf{x}_{i|k-1})^\top \mathbf{S}_i^{-1} (\mathbf{z}_j - \mathbf{H}\mathbf{x}_{i|k-1}),
\end{equation}
where $\mathbf{S}_i = \mathbf{H}\mathbf{P}_{i|k-1}\mathbf{H}^\top + \mathbf{R}$ is the innovation covariance. The Hungarian algorithm is applied to the cost matrix to obtain the globally optimal assignment subject to a gating threshold. Tracks without assigned detections are propagated with increased uncertainty, and tracks remaining unassigned for several consecutive steps are removed.

\section{Distributed Kalman--Consensus Filter with Adaptive Weighting}
\label{sec:dkcf}
We now extend the local tracking framework to a distributed setting in which two robots share information to obtain improved estimates of object states.

\subsection{System Overview}
Consider a network of $N$ robots (in this work, $N = 2$), each performing local detection and tracking as described in Section~\ref{sec:local}. Let $\hat{\mathbf{x}}_i^+(k)$ and $\mathbf{P}_i^+(k)$ denote the locally updated state estimate and covariance maintained by robot $i$ for a given tracked object at time step $k$. The DKCF aims to fuse these local estimates through distributed consensus, while accounting for coordinate frame misalignment and heterogeneous localization uncertainties. The framework builds upon the MOTLEE approach while introducing the proposed adaptive weighting extension.

\subsection{Information-Form Fusion with Frame Alignment}
Building upon the MOTLEE framework, information-form fusion incorporates estimates from neighboring robots. For each neighbor $j \in \mathcal{N}_i$, robot $i$ receives the neighbor's state estimate and covariance, which are transformed into robot $i$'s coordinate frame using relative pose information obtained from the TF tree. For each such estimate, the information vector and matrix are defined as:
\begin{equation}
\mathbf{u}_j = \mathbf{H}^\top \mathbf{R}_j^{-1} \mathbf{z}_j, \quad \mathbf{U}_j = \mathbf{H}^\top \mathbf{R}_j^{-1} \mathbf{H}.
\end{equation}
Aggregating the local and neighbor information:
\begin{equation}
\mathbf{y}_i(k) = \sum_{j \in \mathcal{N}_i \cup \{i\}} \mathbf{u}_j, \quad \mathbf{Y}_i(k) = \sum_{j \in \mathcal{N}_i \cup \{i\}} \mathbf{U}_j.
\end{equation}
The information gain matrix is computed as:
\begin{equation}
\mathbf{M}_i(k) = \left[\mathbf{P}_i^+(k)^{-1} + \mathbf{Y}_i(k)\right]^{-1}.
\end{equation}

The standard DKCF update combines the aggregated information with a consensus term:
\begin{align}
\hat{\mathbf{x}}_i(k) &= \hat{\mathbf{x}}_i^+(k) + \mathbf{M}_i(k)\left[\mathbf{y}_i(k) - \mathbf{Y}_i(k)\hat{\mathbf{x}}_i^+(k)\right] \\
&\quad + \frac{\mathbf{M}_i(k)}{1 + \|\mathbf{M}_i(k)\|} \sum_{j \in \mathcal{N}_i} \left(\hat{\mathbf{x}}_j^+(k) - \hat{\mathbf{x}}_i^+(k)\right),
\end{align}
and the covariance is updated as:
\begin{equation}
\mathbf{P}_i(k) = \mathbf{A}\mathbf{M}_i(k)\mathbf{A}^\top + \mathbf{Q}(k).
\end{equation}
\subsection{Proposed Adaptive Uncertainty Weighting}

Standard consensus assigns equal or fixed weights to all agents. The proposed extension weights each robot's contribution inversely to its localization uncertainty. Let $\sigma_i(k)$ denote the position standard deviation of robot $i$ at time $k$. The adaptive weight is defined as:
\begin{equation}
w_i(k) = \frac{1/\sigma_i(k)}{\sum_{j \in \mathcal{N}_i \cup \{i\}} 1/\sigma_j(k)},
\end{equation}
which normalizes such that $\sum_{j \in \mathcal{N}_i \cup \{i\}} w_j(k) = 1$.

The weighted DKCF update modifies the consensus term:
\begin{align}
\hat{\mathbf{x}}_i^{\text{adapt}}(k) &= \hat{\mathbf{x}}_i^+(k) + \mathbf{M}_i(k)\left[\mathbf{y}_i(k) - \mathbf{Y}_i(k)\hat{\mathbf{x}}_i^+(k)\right] \\
&\quad + \mathbf{M}_i(k) \sum_{j \in \mathcal{N}_i} w_j(k) \left(\hat{\mathbf{x}}_j^+(k) - \hat{\mathbf{x}}_i^+(k)\right).
\end{align}

This formulation gives greater influence to neighbors with better localization accuracy while still maintaining distributed consensus. The adaptive weights respond dynamically to changes in localization uncertainty, which may occur as robots navigate and encounter regions with varying map quality.

\subsection{Frame Alignment and Track Management}
Accurate frame alignment between robots is essential for effective fusion. Each robot maintains TF transforms between its local frames and a common odometry frame. Estimates received from neighbors are first transformed into the odometry frame and then into the local robot frame before fusion. This ensures consistent coordinate systems across the distributed system.

Track identities are maintained through global track IDs and heuristic matching based on spatial proximity and motion consistency. Tracks accumulating large residuals or exhibiting persistent inconsistency with neighbor estimates are flagged as mistracks and removed. This mechanism mitigates the impact of data association errors on the consensus process.

\section{Simulation Results}
\label{sec:results}
The proposed framework was implemented in ROS and evaluated in a Gazebo simulation. Two differential-drive robots equipped with 2D lidars explore a bounded environment containing four cylindrical objects moving along predefined straight-line paths with reversals at boundaries. Maps are built online using the Gmapping package, and all detections and tracks are visualized in RViz.

\subsection{Local Tracking Performance}
\begin{figure}[t]
\centerline{\includegraphics[width=0.9\columnwidth]{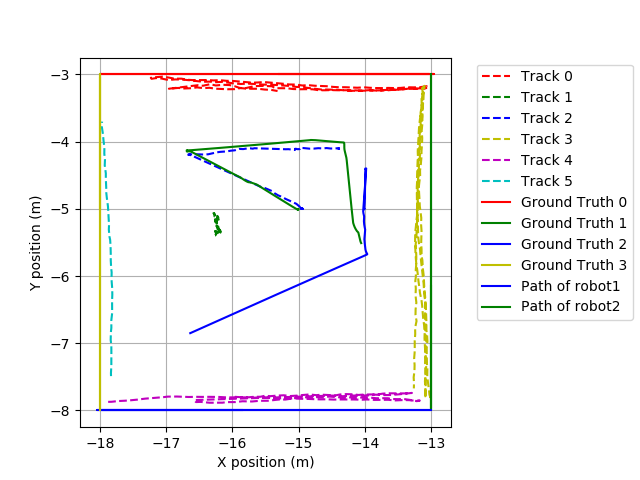}}
\caption{Comparison of estimated tracks (dashed) and ground-truth trajectories (solid) for single-robot KF.}
\label{fig:tracks}
\end{figure}

Figure~\ref{fig:tracks} illustrates estimated tracks versus ground-truth trajectories for the moving objects using single-robot Kalman filtering. Ground truth is shown as solid lines and estimated tracks as dashed lines.

The Kalman Filter with a constant-velocity model successfully tracks most objects, though deviations occur during object reversals and when objects pass in close proximity. DBSCAN occasionally misclassifies static obstacles as moving when spatially adjacent to moving objects, resulting in false positives.

\begin{figure}[t]
\centerline{\includegraphics[width=0.9\columnwidth]{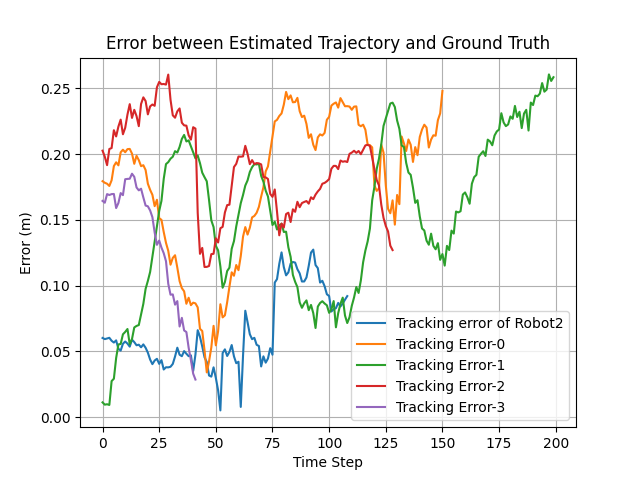}}
\caption{Position error over time for single-robot KF.}
\label{fig:errors_local}
\end{figure}

Figure~\ref{fig:errors_local} displays position error over time. Errors remain relatively small during straight-line motion but exhibit transient spikes during direction reversals and when objects enter or exit the lidar field of view. These characteristics highlight fundamental limitations of constant-velocity models and GNN-based association in highly dynamic scenarios.

\subsection{Quantitative Results}

Table~\ref{tab:mota_stats} summarizes the Multiple Object Tracking Accuracy (MOTA) statistics for both the standard and adaptive consensus configurations. The results demonstrate a distinct divergence in performance based on the specific operating conditions of each agent.

Robot 1, which suffered from higher localization uncertainty during the trials, benefited significantly from the adaptive weighting. The proposed method improved its local MOTA by a mean of \textbf{+0.085} and its global fused estimate by \textbf{+0.093}. This improvement confirms that the adaptive mechanism allowed the agent to prioritize high-confidence neighbor data, effectively anchoring the local estimate against drift-induced misalignment.

Conversely, Robot 2 experienced a decrease in mean MOTA (approximately \textbf{-0.10}). This performance drop suggests that while the adaptive mechanism is robust against noise, it may be overly conservative. By down-weighting neighbor information based on covariance size, the system likely prevented Robot 2 from utilizing valid, helpful data from Robot 1, thereby reducing recall. 

\begin{figure}[t]
\centerline{\includegraphics[width=0.95\columnwidth]{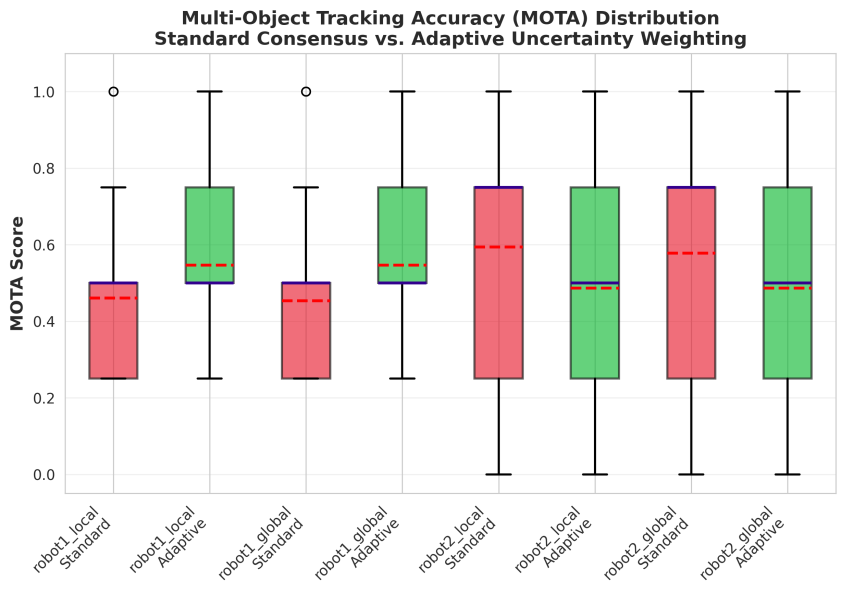}}
\caption{MOTA distribution box plots. The adaptive weighting strategy (right) improves the consistency of Robot 1 by rejecting noise, but reduces the peak performance of Robot 2, highlighting a trade-off between robustness and cooperative gain.}
\label{fig:mota_boxplot}
\end{figure}

Figure~\ref{fig:mota_boxplot} illustrates the distribution of these results. The box plots reveal that while the adaptive approach (bottom) yields a wider interquartile range for Robot 2, it shifts the median performance of the struggling Robot 1 higher. This indicates that the adaptive consensus is particularly effective in heterogeneous networks, prioritizing the stability of agents with poor localization at the cost of reduced cooperative gain for well-localized agents.

\begin{table}[t]
\centering
\caption{MOTA Statistics: Standard Consensus vs. Adaptive Weighting}
\label{tab:mota_stats}
\begin{tabular}{lcccc}
\toprule
\textbf{Configuration} & \textbf{Mean} & \textbf{Median} & \textbf{Std Dev} & \textbf{Range} \\
\midrule
\multicolumn{5}{l}{\textit{Robot 1 (High Uncertainty)}} \\
\midrule
Standard (Local) & 0.461 & 0.500 & 0.189 & [0.25, 1.00] \\
Adaptive (Local) & \textbf{0.546} & 0.500 & 0.216 & [0.25, 1.00] \\
$\Delta$ Mean & \textit{+0.085} & --- & --- & --- \\
\midrule
Standard (Global) & 0.453 & 0.500 & 0.189 & [0.25, 1.00] \\
Adaptive (Global) & \textbf{0.546} & 0.500 & 0.216 & [0.25, 1.00] \\
$\Delta$ Mean & \textit{+0.093} & --- & --- & --- \\
\midrule
\multicolumn{5}{l}{\textit{Robot 2 (Low Uncertainty)}} \\
\midrule
Standard (Local) & 0.594 & 0.750 & 0.325 & [0.00, 1.00] \\
Adaptive (Local) & 0.486 & 0.500 & 0.298 & [0.00, 1.00] \\
$\Delta$ Mean & \textit{-0.108} & --- & --- & --- \\
\midrule
Standard (Global) & 0.578 & 0.750 & 0.323 & [0.00, 1.00] \\
Adaptive (Global) & 0.486 & 0.500 & 0.298 & [0.00, 1.00] \\
$\Delta$ Mean & \textit{-0.092} & --- & --- & --- \\
\bottomrule
\end{tabular}
\end{table}

\subsection{Impact of Localization Uncertainty}

The experimental results highlight an asymmetric response to the adaptive weighting scheme, driven by the disparity in localization fidelity between agents. As evidenced by the baseline local MOTA scores (Table~\ref{tab:mota_stats}), Robot 1 exhibited significantly higher localization uncertainty compared to Robot 2.

Since the adaptive weighting function $\mathcal{W}_{ij}$ scales neighbor influence inversely to the trace of their error covariance, the system effectively enforces a directionality of information flow from lower-uncertainty to higher-uncertainty nodes. This results in two distinct behaviors:

\begin{enumerate}
    \item \textbf{Stabilization of High-Uncertainty Agents:} The agent with higher uncertainty (Robot 1) correctly identifies the lower covariance of its neighbor. By assigning higher weights to these incoming estimates, the struggling agent is able to ``anchor'' its tracks to the more stable reference frame of the neighbor, resulting in a substantial performance gain.
    \item \textbf{Conservative Rejection by Stable Agents:} Conversely, the stable agent (Robot 2) identifies the neighbor's estimates as high-variance. The adaptive mechanism attenuates these inputs to prevent the corruption of its local map. While this successfully preserves estimate purity, it leads to a conservative rejection strategy that slightly reduces global tracking recall by discarding valid, albeit noisy, field-of-view extensions.
\end{enumerate}

This validates the theoretical motivation: uncertainty-aware weighting acts as a ``quality filter,'' significantly improving the stability of worst-case performers while protecting best-case performers from data contamination, albeit at the cost of reduced cooperative gain.

\subsection{Impact of Map Quality and Latency}

\begin{figure}[t]
\centerline{\includegraphics[width=0.9\columnwidth]{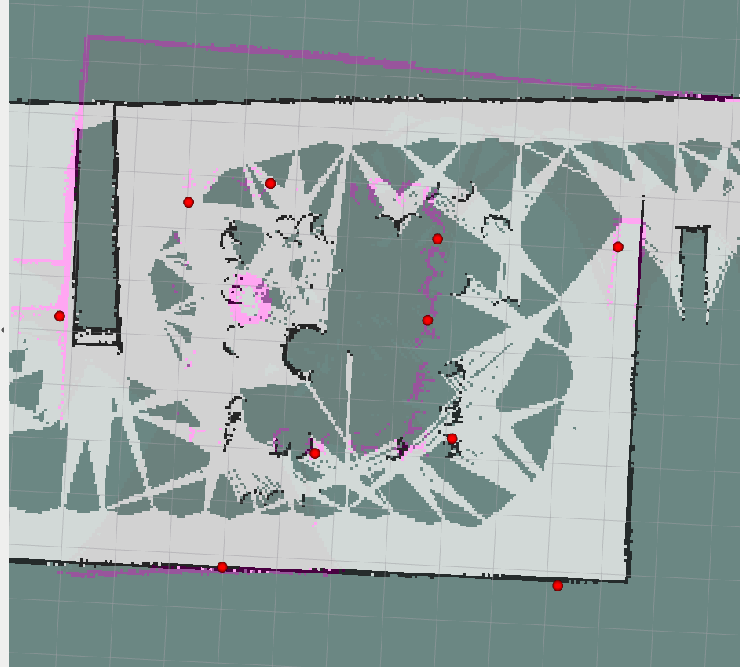}} 
\caption{Real-time Gmapping and object tracking by two robots in a partially constructed map.}
\label{fig:gmapping}
\end{figure}

Distributed tracking accuracy is heavily dependent on the quality of the underlying map and the timeliness of data exchange. Figure~\ref{fig:gmapping} highlights the impact of online SLAM, where unexplored regions introduce localization biases that compromise inter-robot frame alignment. Beyond spatial errors, the asynchronous nature of the network introduces stochastic communication delays. These latencies can cause the fusion of temporally inconsistent estimates, particularly during high-speed maneuvers. A key advantage of the proposed adaptive weighting is its ability to partially compensate for these delays: as latency increases, the uncertainty of the neighbor's estimate grows, causing the consensus mechanism to naturally attenuate its influence.
\subsection{System Limitations}

While the proposed framework enhances robustness, it operates within specific constraints:

\textit{Conservative Rejection Strategy}: The exponential weighting function aggressively prioritizes estimate purity over recall. As observed with the stable agent (Robot 2), this can lead to the rejection of valid data from neighbors with moderate uncertainty, reducing cooperative gain.

\textit{Kinematic Constraints}: The Constant-Velocity model fails to anticipate abrupt maneuvers. Target reversals cause transient innovation spikes that temporarily decouple the consensus fusion until linear motion resumes.

\textit{Dependency on Dynamic Anchors}: The frame alignment module relies on a minimum density of observable objects. In scenarios with sparse dynamic activity or severe map deformation, the relative transform optimization becomes ill-conditioned.

\textit{Residual Latency}: While adaptive weighting attenuates the influence of ``stale'' data, it does not correct the underlying temporal error. Significant communication delays still result in information loss, necessitating predictive time-compensation strategies.

\section{Conclusion}
\label{sec:discussion}
This paper presented an implementation of the MOTLEE framework extended with an adaptive uncertainty weighting mechanism. By dynamically scaling consensus weights based on the trace of neighbor covariance matrices, the proposed method addresses the challenge of heterogeneous localization quality in distributed robot teams.

Experimental results demonstrated that adaptive weighting yields MOTA improvements of approximately \textbf{+0.093} for agents suffering from localization drift. This confirms that the mechanism effectively insulates struggling agents from errors while allowing them to benefit from more stable neighbors. However, a performance trade-off was observed for well-localized agents, where the conservative rejection of noisy neighbor data led to reduced recall.

Future work will focus on: (i) integrating Interacting Multiple-Model (IMM) filters to handle motion reversals, (ii) tighter coupling between SLAM and tracking to reduce map-induced biases, and (iii) developing asymmetric consensus strategies that maximize cooperative gain without compromising local stability.

\section*{Acknowledgment}
The authors acknowledge the financial support provided by the Aero.Next Project under Grant Nos. C645727867 and 00000066.



\begin{thebibliography}{99}

\bibitem{wang2002simultaneous}
C.-C. Wang and C. Thorpe, ``Simultaneous localization and mapping with detection and tracking of moving objects,'' in \textit{Proceedings of the IEEE International Conference on Robotics and Automation (ICRA)}, 2002.

\bibitem{wang2003online}
C.-C. Wang, C. Thorpe, and S. Thrun, ``Online simultaneous localization and mapping with detection and tracking of moving objects: Theory and results from a ground vehicle in crowded urban areas,'' in \textit{Proceedings of the IEEE International Conference on Robotics and Automation (ICRA)}, 2003, pp. 842--849.

\bibitem{wang2007simultaneous}
C.-C. Wang, C. Thorpe, S. Thrun, M. Hebert, and H. Durrant-Whyte, ``Simultaneous localization, mapping and moving object tracking,'' \textit{International Journal of Robotics Research (IJRR)}, vol. 26, no. 9, pp. 889--916, 2007.

\bibitem{choi2012general}
W. Choi, C. Pantofaru, and S. Savarese, ``A general framework for tracking multiple people from a moving camera,'' \textit{IEEE Transactions on Pattern Analysis and Machine Intelligence (TPAMI)}, vol. 35, no. 7, pp. 1577--1591, 2012.

\bibitem{Peterson2023}
M. B. Peterson, P. C. Lusk, and J. P. How, ``MOTLEE: Distributed mobile multi-object tracking with localization error elimination,'' in \textit{2023 IEEE/RSJ International Conference on Intelligent Robots and Systems (IROS)}. IEEE, 2023, pp. 719--726.

\bibitem{tian2022dl}
X. Tian, Z. Zhu, J. Zhao, G. Tian, and C. Ye, ``DL-SLOT: Dynamic lidar SLAM and object tracking based on collaborative graph optimization,'' \textit{arXiv preprint arXiv:2212.02077}, 2022.

\bibitem{ref33}
C.-C. Wang and C. Thorpe, ``Simultaneous localization and mapping with detection and tracking of moving objects,'' in \textit{Proceedings of the IEEE International Conference on Robotics and Automation (ICRA)}, 2002.

\bibitem{ref37}
F. Dellaert and M. Kaess, ``Factor graphs for robot perception,'' \textit{Foundations and Trends in Robotics}, vol. 4, no. 1--2, pp. 1--139, 2017.

\bibitem{ref38}
H. Zhang, H. Uchiyama, S. Ono, and H. Kawasaki, ``MOTSLAM: MOT-assisted monocular dynamic SLAM using single-view depth estimation,'' in \textit{Proceedings of the IEEE/RSJ International Conference on Intelligent Robots and Systems (IROS)}, 2022, pp. 4865--4872.

\bibitem{lin2004exact}
X. Lin, Y. Bar-Shalom, and T. Kirubarajan, ``Exact multisensor dynamic bias estimation with local tracks,'' \textit{IEEE Transactions on Aerospace and Electronic Systems}, vol. 40, no. 2, pp. 576--590, 2004.

\bibitem{moratuwage2013collaborative}
D. Moratuwage, B.-N. Vo, and D. Wang, ``Collaborative multi-vehicle SLAM with moving object tracking,'' in \textit{2013 IEEE International Conference on Robotics and Automation}. IEEE, 2013, pp. 5702--5708.

\bibitem{ahmad2013cooperative}
A. Ahmad, G. D. Tipaldi, P. Lima, and W. Burgard, ``Cooperative robot localization and target tracking based on least squares minimization,'' in \textit{2013 IEEE International Conference on Robotics and Automation}. IEEE, 2013, pp. 5696--5701.

\bibitem{ref1}
M. B. Peterson, P. C. Lusk, and J. P. How, ``MOTLEE: Distributed mobile multi-object tracking with localization error elimination,'' in \textit{2023 IEEE/RSJ International Conference on Intelligent Robots and Systems (IROS)}. IEEE, 2023, pp. 719--726.

\bibitem{olfati2007consensus}
R. Olfati-Saber, J. A. Fax, and R. M. Murray, ``Consensus and cooperation in networked multi-agent systems,'' \textit{Proceedings of the IEEE}, vol. 95, no. 1, pp. 215--233, 2007.

\end{thebibliography}
\end{document}